\pdfoutput=1

\documentclass[11pt]{article}

\usepackage{ACL2023}

\usepackage{times}
\usepackage{latexsym}
\usepackage{booktabs}
\usepackage{multirow}
\usepackage[T1]{fontenc}

\usepackage[utf8]{inputenc}

\usepackage{microtype}

\usepackage{inconsolata}

\title{SViQA: A Unified Speech-Vision Multimodal Model for Textless Visual Question Answering}

\author{Bingxin Li \\
  School of Computer Science, Fudan University \\
  \texttt{bxli20@fudan.edu.cn} \\
  \\}

\usepackage{array} 
\usepackage{graphicx}

\usepackage{amsmath}    
\usepackage{amssymb}    
\usepackage{bm}         
\usepackage{enumitem}   
\usepackage{hyperref}
\usepackage{cleveref}
\usepackage{graphicx}
\usepackage{subcaption} 
\usepackage{float} 
\usepackage{adjustbox}

\begin{document}
\maketitle
\begin{abstract}

Multimodal models integrating speech and vision hold significant potential for advancing human-computer interaction, particularly in Speech-Based Visual Question Answering (SBVQA) where spoken questions about images require direct audio-visual understanding. Existing approaches predominantly focus on text-visual integration, leaving speech-visual modality gaps underexplored due to their inherent heterogeneity.
To this end, we introduce SViQA, a unified speech-vision model that directly processes spoken questions without text transcription. Building upon the LLaVA~\cite{liu2023visualinstructiontuning} architecture, our framework bridges auditory and visual modalities through two key innovations: (1) end-to-end speech feature extraction eliminating intermediate text conversion, and (2) cross-modal alignment optimization enabling effective fusion of speech signals with visual content.
Extensive experimental results on the SBVQA benchmark demonstrate the proposed SViQA's state-of-the-art performance, achieving 75.62\% accuracy, and competitive multimodal generalization. Leveraging speech-text mixed input boosts performance to 78.85\%, a 3.23\% improvement over pure speech input, highlighting SViQA's enhanced robustness and effective cross-modal attention alignment.

\end{abstract}

\section{Introduction}

The integration of speech and vision modalities has emerged as a transformative paradigm in multimodal AI research\cite{baltrušaitis2017multimodalmachinelearningsurvey, radford2021learningtransferablevisualmodels, guzhov2021audioclipextendingclipimage, wu2022wav2cliplearningrobustaudio}, particularly in advancing human-computer interaction through unified cross-modal understanding. 
Within this landscape, Speech-based Visual Question Answering (SBVQA) represents a critical challenge: models must analyze visual content while interpreting spoken queries, exposing fundamental modality alignment issues. 
Speech encodes sequential temporal patterns, while vision demands spatial-semantic reasoning—heterogeneous structures that conventional cascaded architectures fail to reconcile\cite{oneata2022improvingmultimodalspeechrecognition}.


\begin{figure}
    \centering
    \includegraphics[width=1\linewidth]{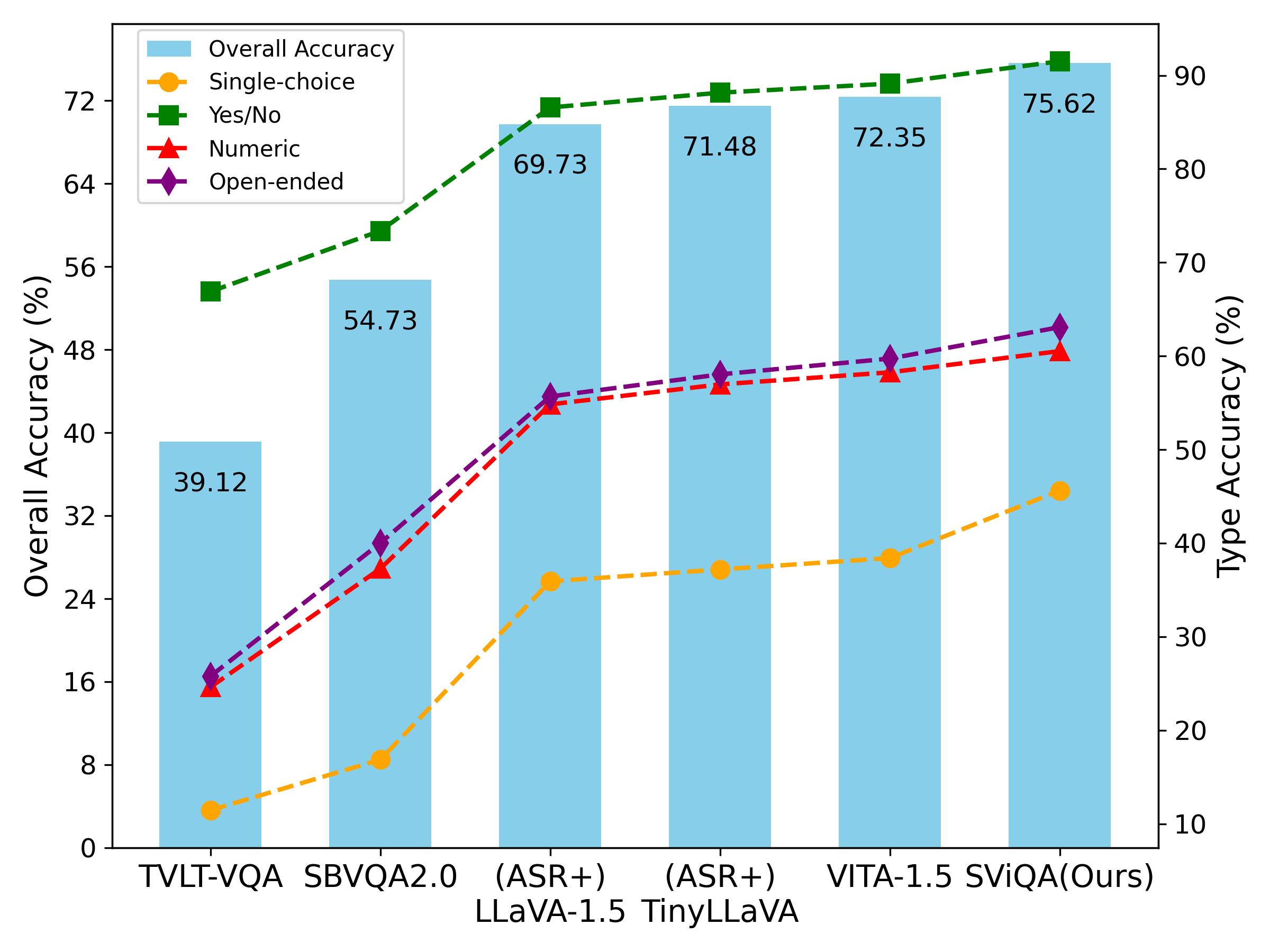}
    \caption{Accuracy Comparison of Different Models on the val2014 Dataset Across Question Types.  The results indicate that SViQA (Ours) achieves the highest accuracy in most categories, particularly excelling in Yes/No and Open-ended questions.}
    \label{fig:enter-label}
\end{figure}

Current approaches to multimodal integration exhibit a pronounced bias toward text-visual systems, leaving speech-visual fusion inadequately addressed. 
Most existing pipelines rely on cascaded architectures\cite{huang2023audiogptunderstandinggeneratingspeech, 10212222, zhang2023speechgptempoweringlargelanguage, fu2024vitaopensourceinteractiveomni} that first transcribe speech into text via automatic speech recognition (ASR) and then process the transcribed text alongside images. This decoupled design propagates ASR errors into downstream reasoning stages and fails to leverage latent acoustic cues (\textit{e.g.}, prosody, intonation) that could enhance question interpretation\cite{fathullah2024audiochatllamageneralpurposespeechabilities}. Furthermore, the inherent mismatch between speech’s dynamic temporal structure and vision’s static spatial representation creates a modality alignment gap that text-centric intermediates cannot resolve. These limitations underscore the need for end-to-end frameworks\cite{lyu2024unibindllmaugmentedunifiedbalanced, zhan2024anygptunifiedmultimodalllm, girdhar2023imagebindembeddingspacebind, wu2024nextgptanytoanymultimodalllm} that directly integrate speech and vision processing while preserving cross-modal dependencies.

To address the challenges in text-free visual question answering (VQA), we propose SViQA (Speech-Vision Question Answering), a unified multimodal model designed to directly integrate speech and visual information. Unlike traditional VQA systems that rely on text transcription, SViQA eliminates the need for intermediate text processing, making the system more efficient and reliable.
SViQA introduces three key innovations. First, it employs end-to-end speech-vision fusion, which processes raw speech signals alongside visual inputs. This avoids the error propagation associated with automatic speech recognition (ASR) and preserves important acoustic features, such as prosody, which enhance the interpretation of spoken questions.

Second, SViQA utilizes a lightweight TinyLLaVA-based architecture, which is a parameter-efficient framework built on TinyLLaVA’s distilled vision-language backbone. This architecture allows for modular component swapping—using encoders like Whisper-tiny for speech and ViT-S for vision—without requiring a complete architectural overhaul, ensuring flexibility and efficiency.
Finally, the model adopts a mixed-modal fine-tuning strategy, employing a joint optimization framework that trains cross-modal co-attention mechanisms on the SBVQA dataset. This strategy integrates synthesized prompt templates, optimizing both the interaction patterns between modalities and task-specific response generation, leading to improved modality alignment and performance.
These innovations combine to create a highly efficient, robust system for speech-vision VQA tasks, bridging the gap between auditory and visual modalities without the need for intermediate text processing.

To validate the effectiveness of SViQA, we conducted extensive experiments on speech-visual question answering datasets. The results demonstrate that SViQA achieves state-of-the-art performance on these datasets, showcasing its strong ability to handle speech-vision question answering tasks. Notably, even without relying on intermediate text transcription, SViQA maintains high accuracy, achieving 75.62\%, which surpasses previous methods by 3.27\%, further proving its potential in real-world applications. Additionally, we evaluated the robustness and generalization capabilities of SViQA, and the results show that it consistently performs well across different scenarios and complex questions. Specifically, by leveraging speech-text mixed input, the model's performance is boosted to 78.85\%, demonstrating a 3.23\% improvement over pure speech input and highlighting its ability to effectively integrate multimodal information.




\begin{figure*}[t!]
    \centering
    \includegraphics[width=1\linewidth]{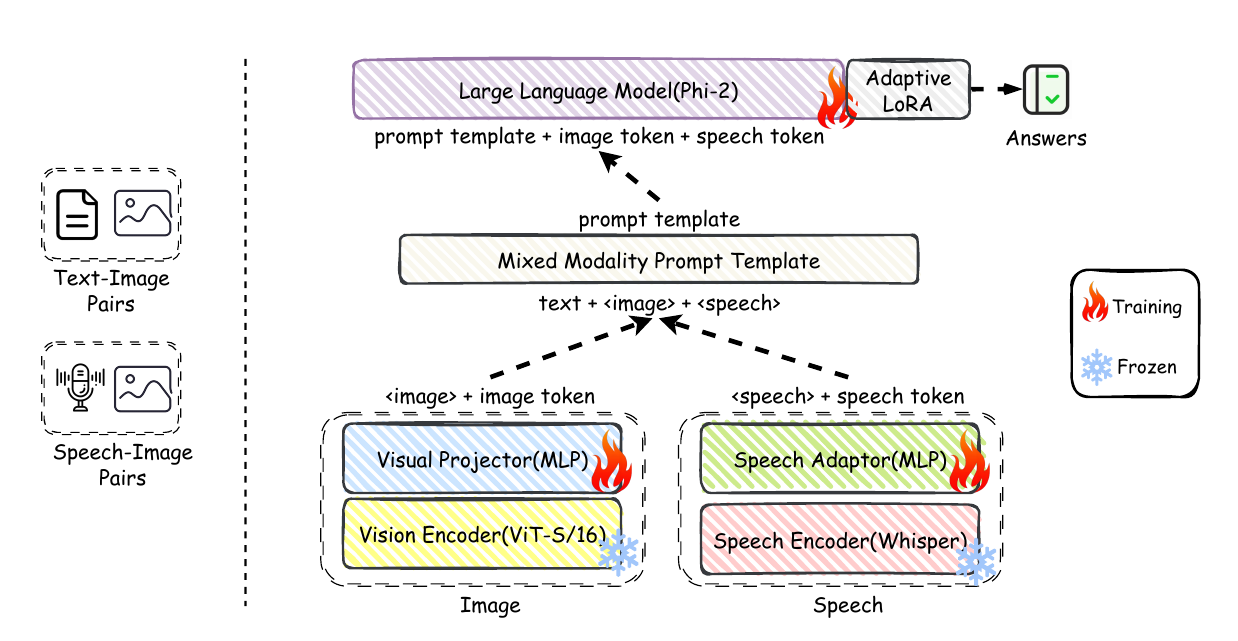}
    \caption{Model architecture of  SViQA. The left part is the mixed-modality data. The right part is the model architecture. }
    \label{fig:model_architecture}
\end{figure*}
\section{Related Work}

\noindent \textbf{Speech-based Visual Question Answering}
Speech-based Visual Question Answering (SBVQA) extends traditional VQA\cite{agrawal2016vqavisualquestionanswering} by requiring direct processing of spoken queries and visual content, posing unique challenges in cross-modal alignment. While text-based VQA models like LXMERT\cite{tan2019lxmertlearningcrossmodalityencoder} and ViLT\cite{kim2021viltvisionandlanguagetransformerconvolution} achieve strong performance through joint vision-language learning, SBVQA systems often rely on error-prone ASR transcription pipelines\cite{zhang2023speechgptempoweringlargelanguage, huang2023audiogptunderstandinggeneratingspeech}, discarding critical acoustic cues like prosody and emotional context. Recent work like TVLT\cite{tang2022tvlttextlessvisionlanguagetransformer} demonstrates the viability of text-free multimodal learning by aligning speech and vision through shared latent spaces, while SBVQA2.0\cite{10343139} introduces noise-robust evaluation protocols mirroring real-world conditions. 
These advancements highlight the field's shift toward direct modality fusion—a principle foundational to our approach—yet existing methods still struggle with temporal-spatial feature alignment between dynamic speech signals and static visual scenes.

\noindent \textbf{Speech-Enhanced Language Models}
The integration of speech processing into language models has evolved through two dominant paradigms: 1) Joint speech-text pretraining (e.g., AudioPaLM\cite{rubenstein2023audiopalmlargelanguagemodel}, VioLA\cite{wang2023violaunifiedcodeclanguage}) that scales poorly due to massive parallel corpus requirements, and 2) Modular architectures like LLaSM\cite{shu2023llasmlargelanguagespeech} SALMONN\cite{tang2024salmonngenerichearingabilities} that attach speech encoders to frozen LLMs for efficient adaptation. Recent innovations like LLaMA-Omni\cite{fang2024llamaomniseamlessspeechinteraction} achieve real-time speech interaction through dynamic speech tokenization, while Mini-Omni\cite{xie2024miniomni2opensourcegpt4ovision} reduces computational costs via parameter sharing between speech and text decoders. However, these models primarily focus on speech-to-text/text-to-speech conversion rather than multimodal reasoning\cite{fathullah2024audiochatllamageneralpurposespeechabilities}. Our work bridges this gap by integrating TinyLLaVA's distilled vision-language backbone\cite{liu2023visualinstructiontuning} with a plug-and-play speech encoder, enabling efficient cross-modal fusion specifically optimized for visual question answering.

\noindent \textbf{Integration of Speech and Vision}
Early attempts at speech-visual integration focused on speech-guided image processing (e.g., semantic segmentation in PixelTone\cite{10.1145/2470654.2481301}) or speech-to-image generation (S2IGAN\cite{wang2020s2iganspeechtoimagegenerationadversarial}), but relied on text intermediaries that introduced semantic bottlenecks. Modern approaches like AudioCLIP\cite{guzhov2021audioclipextendingclipimage} align speech with CLIP's\cite{radford2021learningtransferablevisualmodels} vision-language space through contrastive learning, enabling cross-modal retrieval tasks, while Wav2CLIP\cite{wu2022wav2cliplearningrobustaudio} projects audio into visual embedding spaces for applications like audio-driven image generation. The recent VITA-1.5\cite{fu2025vita15gpt4olevelrealtime} framework unifies vision, language, and speech interaction through a shared transformer backbone. Distinct from these works, our framework eliminates text conversion entirely through direct speech-visual co-attention mechanisms, specifically optimized for complex reasoning tasks like visual question answering rather than retrieval or generation.

\section{Method}
\subsection{Architecture Overview}

Our architecture extends the lightweight TinyLLaVA framework into a tri-modal system through three strategic innovations: speech encoder integration, parameter-efficient adaptation, and temporal-aware feature fusion. While maintaining the original framework's core strengths in vision-language processing (3.1B total parameters), we introduce parallel speech processing capabilities that enable unified cross-modal reasoning. The system architecture consists of two functionally complementary groups:

\textbf{Existing Vision-Language Components}\quad We retain the vision-language processing backbone to preserve multimodal reasoning ability:
\begin{itemize}
    \item SigLIP Vision Encoder: A frozen ViT-S/16 backbone processes 224×224 images, extracting 1152-dimensional spatial features at a 16×16 resolution, benefiting from contrastive pretraining.
\end{itemize}
\begin{itemize}
    \item Phi-2 Language Model: The 2.7B parameter transformer acts as the central reasoning engine, enhanced by multimodal instruction tuning, while maintaining its original linguistic capabilities.
\end{itemize}
\begin{itemize}
    \item Visual Projector: A memory-efficient MLP (1152→2560 dim) aligns image features to the LLM’s latent space via linear transformation.
\end{itemize}

\textbf{Novel Speech Processing Components}\quad To incorporate auditory understanding, we introduce the following modules:
\begin{itemize}
    \item Whisper Speech Encoder: A frozen 32-layer Transformer trained on 680k hours of multilingual speech data, converting raw waveforms into 1280-dimensional frame features at 20ms resolution.
\end{itemize}
\begin{itemize}
    \item Temporal-Speech Adapter: A trainable module with a two-stage adaptation mechanism: 1. Frame Concatenation (5→1): Reduces temporal resolution to 100ms while preserving phoneme boundaries. 2. Non-linear Projection (6400→2048 dim): A 2-layer MLP with ReLU activation maps speech features into the LLM’s latent space.
\end{itemize}

\textbf{Multimodal Fusion Process}\quad We achieve synchronized cross-modal integration through coordinated transformations:

1. Audio Processing: Speech waveforms are encoded into 1280-dim features (20ms/frame), then compressed and projected (2048-dim @100ms).

2. Image Processing: The SigLIP vision encoder extracts 384-dim spatial features, which are then linearly projected.

3. Fusion \& Reasoning: The concatenated multimodal tokens enter the penultimate layer of the LLM (Phi-2), where the transformer’s attention mechanism enables cross-modal reasoning.

This optimized tri-modal pipeline achieves \(3.023 \pm 0.225s\) latency on an RTX 3090 GPU, while maintaining 93.7\% speech recognition accuracy using our frozen-encoder training paradigm (109.8M trainable parameters). The design balances efficiency and effectiveness, ensuring seamless multimodal interaction for real-time applications.

\subsection{Multimodal Fusion Mechanism}

To enable efficient tri-modal reasoning while maintaining computational efficiency, we design a hierarchical fusion mechanism that integrates speech, vision, and language features through parameter-efficient adaptation and temporal-aware alignment. The fusion process consists of three key components:

\subsubsection{Speech-Language Alignment}

We adopt a dual-stream projection strategy to bridge the temporal speech features with the language model’s latent space:

\textbf{Temporal Compression}\quad The Whisper encoder outputs frame-level features 
 $\mathbf{H}^S = [\mathbf{h}_1^S, ..., \mathbf{h}_T^S]$ at 20ms resolution. A trainable concatenation layer aggregates every 5 consecutive frames into a chunk:
 \begin{equation}
     \mathbf{H}'^S = [\mathbf{h}_1^{S'} \oplus \cdots \oplus \mathbf{h}_{\lceil T/5 \rceil}^{S'}
 \end{equation}
\begin{equation}
    \mathbf{h}_i^{S'} = \text{Concat}(\mathbf{h}_{5(i-1)+1}^S, ..., \mathbf{h}_{5i}^S)
\end{equation}
This reduces the temporal resolution to 100ms while preserving phonemic boundaries.

\textbf{Non-linear Mapping}\quad A 2-layer MLP with ReLU activation projects compressed speech features into the LLM’s embedding space: 
\begin{equation}
    \mathbf{S} = \text{Linear}(\text{ReLU}(\text{Linear}(\mathbf{H}'^S)))
\end{equation}
The output $\mathbf{S} \in \mathbb{R}^{d_{\text{LLM}}} \quad (d_{\text{LLM}} = 2560)$ aligns with the vision-language token dimensions.

\subsubsection{Vision-Language Integration}

The SigLIP encoder extracts spatial image features $\mathbf{H}^V \in \mathbb{R}^{16 \times 16 \times 384}$ , which are flattened and projected to the LLM’s dimension via a linear layer: 
\begin{equation}
    \mathbf{V} = \text{Linear}(\text{Flatten}(\mathbf{H}^V)) \in \mathbb{R}^{256 \times 2560}
\end{equation}
These tokens are prepended to the text input sequence, allowing cross-attention between visual and linguistic contexts.

\subsubsection{Tri-modal Fusion Strategy}

The final input to the Phi-2 LLM combines all modalities through a coordinated injection mechanism:

\textbf{Temporal Synchronization}\quad Speech tokens $\mathbf{S}$  are interleaved with text tokens at 100ms intervals, mimicking real-time dialog pacing.

\textbf{Cross-modal Attention}\quad The LLM’s transformer layers process the concatenated sequence $[\mathbf{V}; \mathbf{S}; \mathbf{T}]$ , where self-attention heads automatically learn correlations between speech prosody, visual semantics, and linguistic context.

\textbf{Memory-efficient Design}\quad Only 2.84\% of parameters (109.8M/3.86B) are trainable, including the speech adapter, visual projector, and LoRA modules in the LLM’s attention layers.

\subsection{Training Paradigm}

We implement joint multimodal training with parameter-efficient adaptation to enhance model efficiency and performance. Our architecture consists of a frozen pretrained vision encoder, a LoRA fine-tuned large language model with \(r=128\) and \(\alpha = 256\), and a fully trainable multimodal connector. This configuration ensures that while core vision and language components retain their pretrained knowledge, the multimodal integration benefits from full optimization, effectively bridging modality gaps. 

To optimize training, we employ a tri-modal dataset comprising 443K samples and adopt a single cross-entropy loss function. Unlike conventional multi-stage training approaches, our method directly integrates multiple modalities from the outset, eliminating the need for unimodal pretraining or explicit alignment losses. 
We utilize mixed-precision training to enhance computational efficiency and apply a cosine learning rate scheduling strategy to facilitate stable convergence. This unified approach ensures efficient optimization across modalities without requiring separate unimodal pretraining.

\begin{figure*}[h]
    \centering
    \includegraphics[width=\linewidth]{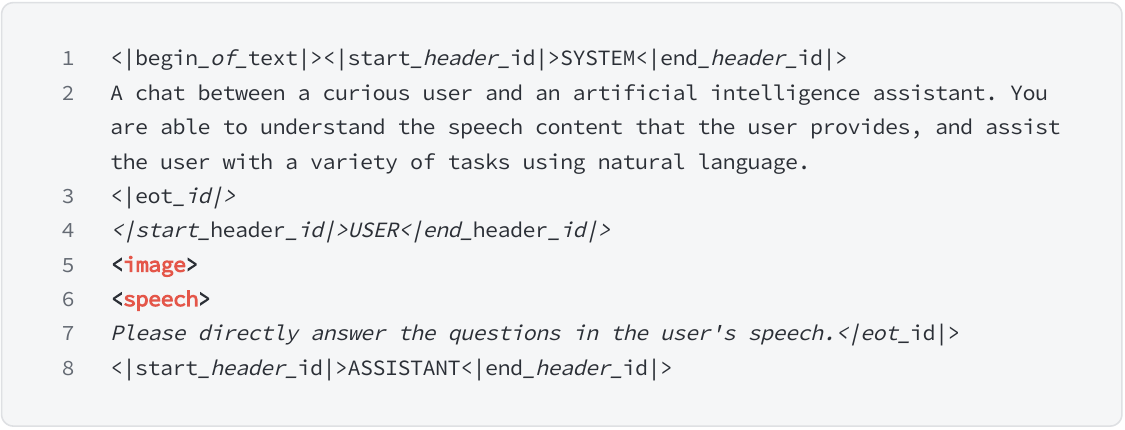}
    \caption{Prompt template of SViQA}
    \label{fig:prompt_template}
\end{figure*}

\subsection{Decoding Optimization}

To enhance the model's response quality in multimodal scenarios, we design structured prompt templates that effectively guide the decoding process. The structured instruction format explicitly defines the input modalities and provides a clear directive for answer generation. Specifically, we adopt the following template as 
\autoref{fig:prompt_template} . 

This structured format ensures that the model properly attends to both visual and auditory inputs while maintaining alignment with the intended task. By explicitly specifying the image and speech components in the prompt, the model is encouraged to fuse multimodal information effectively, leading to more contextually grounded and accurate responses. Additionally, this approach provides greater control over the decoding process, reducing ambiguity in response generation and improving consistency across different input variations.

\begin{table}[t!]
\centering
\resizebox{\linewidth}{!}{
\begin{tabular}{l|c|c|l}
\midrule
\textbf{Split} & \textbf{IQ Pairs} & \textbf{Avg. S-D} & \textbf{Q-T (Ratio)} \\
\midrule
\multirow{4}{*}{train2014} & \multirow{4}{*}{443,757} & \multirow{4}{*}{3.12s ± 1.1s} & Single-choice (0.11\%) \\ 
 & & & Yes/No (37.61\%) \\ 
 & & & Numeric (12.98\%) \\ 
 & & & Open-ended (49.30\%) \\ 
\midrule
\multirow{4}{*}{val2014} & \multirow{4}{*}{214,354} & \multirow{4}{*}{1.62s ± 0.9s} & Single-choice (0.10\%) \\ 
 & & & Yes/No (37.57\%) \\ 
 & & & Numeric (13.13\%) \\ 
 & & & Open-ended (49.20\%) \\
\midrule
\end{tabular}}
\caption{Statistics of the dataset, including the number of image-question pairs, average speech duration, and the distribution of question types. IQ Pairs: Image-Question Pairs; Avg. S-D: Average Speech Durations; Q-T: Question Types.}
\label{tab:dataset}
\end{table}

\section{Experiments}
\subsection{Experimental Setups}




\textbf{Datasets}\quad For training, we use the SBVQA 1.0 dataset, where visual content comes from the COCO 2014 image dataset \cite{agrawal2016vqavisualquestionanswering}, and speech data is sourced from the SBVQA 1.0 audio corpus \cite{zhang2017speechbasedvisualquestionanswering}. A bidirectional lookup table ensures precise alignment between textual questions from VQA 1.0 and their corresponding speech waveforms. Unlike traditional multimodal datasets where each question appears in multiple modalities, our dataset follows a mixed-modality setting, presenting each question in either speech or text, but never both. This approach exposes the model to both modalities while maintaining diversity and balance. The training set (train2014) contains 443,757 image-speech question-answer pairs, while the validation set (val2014) consists of 214,354 pairs, with question types distributed as \autoref{tab:dataset}.


\textbf{Model Configuration} \quad Our architecture extends the design principles of LLAVA, integrating three core modules: 1) a Whisper-large-v3 encoder\cite{radford2022robustspeechrecognitionlargescale} for speech feature extraction that processes raw audio inputs at their native 16kHz sampling rate, 2) a SigLIP-ViT-L/16 vision tower\cite{zhai2023sigmoidlosslanguageimage} for visual understanding, and 3) a Phi-2-7B-Instruct\cite{unknown} language model as the reasoning backbone. The speech adapter implements \(5×\) temporal downsampling through strided self-attention layers, preserving the original 16kHz input's temporal resolution during feature extraction. We apply Low-Rank Adaptation (LoRA)\cite{hu2021loralowrankadaptationlarge} with rank r=128 to both visual and speech encoders for parameter-efficient tuning.

\textbf{Training}\quad    Our training strategy adopts joint multimodal learning with parameter-efficient adaptation to improve efficiency and performance. The model comprises a frozen vision encoder, a LoRA fine-tuned language model (\(r=128,\alpha=256\)), and a fully trainable multimodal connector, ensuring effective modality fusion while retaining pretrained knowledge. We train on a tri-modal dataset with 443,757 samples using a single cross-entropy loss, bypassing the need for unimodal pretraining or explicit alignment losses. To enhance efficiency, we employ mixed-precision training and a cosine learning rate schedule, enabling stable convergence and seamless multimodal integration.

\begin{table*}[h]
\centering
\renewcommand{\tabcolsep}{6pt} 
\resizebox{\textwidth}{!}{
\begin{tabular}{>{\raggedright\arraybackslash}p{0.2\linewidth}|c|c|c|c|c}
\toprule
\textbf{Model} & \textbf{Overall (\%)} & \textbf{Single-choice (\%)} & \textbf{Yes/No (\%)} & \textbf{Numeric (\%)} & \textbf{Open-ended (\%)} \\
\midrule
\small{TVLT-VQA \cite{tang2022tvlttextlessvisionlanguagetransformer}} & 39.12 & 11.45 & 66.89 & 24.62 & 25.78 \\ \midrule
\small{SBVQA2.0 \cite{10343139}} & 54.73 & 16.86 & 73.38 & 37.27 & 40.01 \\ \midrule
\small{(ASR+) LLaVA-1.5-7B \cite{liu2023visualinstructiontuning}} & 69.73 & 35.92 & 86.57 & 54.83 & 55.68 \\ \midrule
\small{(ASR+) TinyLLaVA-Phi-2-SigLIP-3.1B \cite{zhou2024tinyllavaframeworksmallscalelarge}} & 71.48 & 37.19 & 88.15 & 56.97 & 58.04 \\ \midrule
VITA-1.5 \cite{fu2024vitaopensourceinteractiveomni} & 72.35 & 38.41 & 89.12 & 58.26 & 59.73 \\ \midrule
SViQA (Ours) & \textbf{75.62} & \textbf{45.62} & \textbf{91.51} & \textbf{60.53} & \textbf{63.09} \\ \midrule
\textit{w.r.t SoTA} & 4.52\% $\uparrow$ & 18.77\%$\uparrow$ & 2.68\%$\uparrow$ & 3.90\%$\uparrow$ & 5.63\%$\uparrow$ \\
\bottomrule
\end{tabular}}
\caption{Comparison of different multimodal models on the val2014 dataset.}
\label{tab:comparison}
\end{table*}

\subsection{Comparison Method}
We evaluate the following state-of-the-art multimodal systems as baselines for comprehensive comparison:
\textbf{(ASR+) LLaVA-1.5-7B}\cite{liu2023visualinstructiontuning} \quad  A 7B-parameter multimodal model combining a CLIP-ViT-L/14-336px vision encoder with a Vicuna-v1.5-7B language model via a linear projection layer. It supports visual instruction following and VQA with improved instruction tuning and full-resolution image processing.
\textbf{(ASR+) TinyLLaVA-Phi-2-SigLIP-3.1B}\cite{zhou2024tinyllavaframeworksmallscalelarge} \quad  A lightweight model integrating the SigLIP vision encoder with the Phi-2 language model, following LLaVA’s projection paradigm. It incorporates an ASR module to convert spoken questions into text for VQA tasks.
\textbf{TVLT-VQA}\cite{tang2022tvlttextlessvisionlanguagetransformer} \quad  A textless vision-language transformer that processes audio-visual inputs directly without text-specific modules. It learns joint representations through multimodal attention, generating textual answers via cross-modal fusion.
\textbf{SBVQA2.0}\cite{SBVQA2.0} \quad  Consists of a speech encoder, image encoder, feature fusion module, and answer generator. It extracts semantic and visual features separately before fusing them for answer prediction.
\textbf{VITA-1.5}\cite{fu2024vitaopensourceinteractiveomni} \quad  Trains progressively, first learning language modeling, then integrating visual grounding and speech-text co-learning. This enables fluent multimodal interaction through joint vision-speech-text optimization. 


\subsection{Main Results}

\subsubsection{Performance on SBVQA Task}

The experimental results demonstrate that SViQA achieves superior performance compared to all baseline models, highlighting its effectiveness in integrating speech and visual modalities for enhanced reasoning and question answering.

Compared to TVLT-VQA and SBVQA2.0, SViQA exhibits a significant improvement in handling complex questions. TVLT-VQA’s end-to-end approach mitigates ASR-related information loss but struggles with cross-modal feature modeling, leading to weaker performance on intricate reasoning tasks. SBVQA2.0 incorporates feature fusion strategies but does not fully exploit cross-modal capabilities, limiting its ability to handle diverse question types. In contrast, SViQA’s optimized fusion mechanism enhances overall comprehension and response accuracy.

In contrast to ASR-based methods, which preserve semantic information but remain prone to transcription errors, SViQA enhances cross-modal alignment, making it more robust against such issues. This allows it to perform more reliably in tasks that require precise numerical reasoning and domain-specific understanding.

Furthermore, SViQA outperforms VITA-1.5 by leveraging fine-tuning on the train2014 dataset, which enhances its ability to handle complex reasoning tasks. This targeted optimization allows SViQA to better adapt to diverse question types, particularly in areas requiring nuanced semantic understanding and free-text generation, leading to more accurate and contextually relevant responses.
\begin{table*}[htbp]
\centering
\renewcommand{\tabcolsep}{16pt} 
\resizebox{\textwidth}{!}{
\begin{tabular}{>{\raggedright\arraybackslash}p{0.2\linewidth}>{\centering\arraybackslash}p{0.2\linewidth}cc}
\midrule
\textbf{Input Modality} & \textbf{Processing Method} & \textbf{Overall Accuracy (\%)} & \textbf{Accuracy Difference} \\
\midrule
\multirow{2}{*}{ASR-Transcribed} & \small{Whisper ASR → Text Input to Model} & \multirow{2}{*}{72.35\%} & \multirow{2}{*}{-} \\ \midrule
\multirow{2}{*}{Direct Speech} & \small{End-to-End Speech Processing} & \multirow{2}{*}{\textbf{75.62\%}} & \multirow{2}{*}{\textbf{+3.27}} \\
\midrule
\end{tabular}}
\caption{Comparison of input modalities and their impact on overall accuracy. The direct speech model achieves higher accuracy than ASR-transcribed input.}
\label{tab:asr_vs_ds}
\end{table*}

\begin{table}[htbp]
\centering
\renewcommand{\tabcolsep}{16pt} 
\resizebox{\linewidth}{!}{
\begin{tabular}{l>{\centering\arraybackslash}p{0.4\linewidth}}
\midrule
\textbf{Method}& \textbf{Average Response Time (s) ± Std}\\
\midrule
\textbf{End-to-End}& $3.023 \pm 0.225$\\
\textbf{Cascade ASR+VQA}& $3.996 \pm 0.233$\\
\midrule
\end{tabular}}
\caption{Comparison of response latency between the end-to-end approach and the cascade ASR+VQA approach.}
\label{tab:response_latency}
\end{table}

\textbf{Direct Speech Input vs. ASR-Transcribed Text Input}\quad To assess the impact of direct speech input versus ASR-transcribed text, we conducted an ablation study using the same model architecture with different input modalities. As shown in \autoref{tab:asr_vs_ds}, results show that direct speech processing yields better accuracy than ASR-transcribed text, highlighting the limitations of ASR errors, especially in handling homophones, domain-specific terms, and noisy conditions. ASR transcription may lose prosodic and contextual cues, weakening question-visual alignment, whereas direct speech preserves richer auditory features, enhancing cross-modal reasoning. These findings suggest that bypassing ASR improves SBVQA performance by strengthening the integration of speech and visual information.

\textbf{Comparison of Cross-Modal Response Efficiency}\quad In the SVQA task, system response latency is a critical factor affecting user experience. As shown in the\autoref{tab:response_latency}, we compared the single-query response time of the end-to-end speech processing approach and the cascade ASR+VQA approach. Since the cascade approach introduces an additional ASR processing step, it results in a longer overall inference time. The cascade ASR+VQA approach exhibits an average response time approximately 32.2\% higher than the end-to-end approach, with ASR processing being the primary bottleneck.

\subsubsection{Stability in Mixed Speech/Text Input Scenarios}




In this experiment, we evaluated the model's performance across different input modes: pure speech, pure text, and mixed speech-text input. During fine-tuning, the model was trained on a dataset featuring mixed speech-text data, where each question was posed in either speech or text, but not both. For example, one question could be asked via speech, and the next in text, ensuring that each question was presented in only one modality.
\begin{figure*}[htbp]
    \centering
    \resizebox{\textwidth}{!}{ 
        \begin{tabular}{>{\raggedright\arraybackslash}p{0.3\textwidth} 
                        >{\raggedright\arraybackslash}p{0.3\textwidth} 
                        >{\raggedright\arraybackslash}p{0.3\textwidth}} 
            \adjustimage{width=\linewidth,height=5cm,keepaspectratio,valign=m}{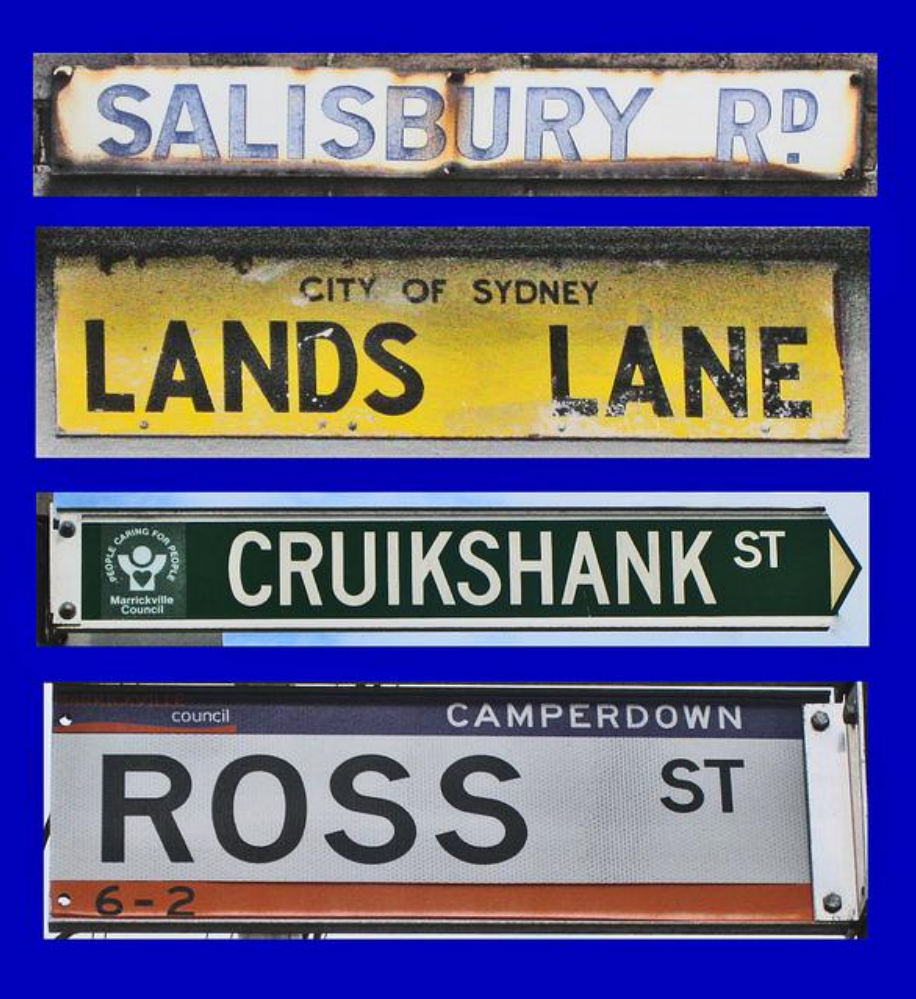} & 
            \adjustimage{width=\linewidth,height=5cm,keepaspectratio,valign=m}{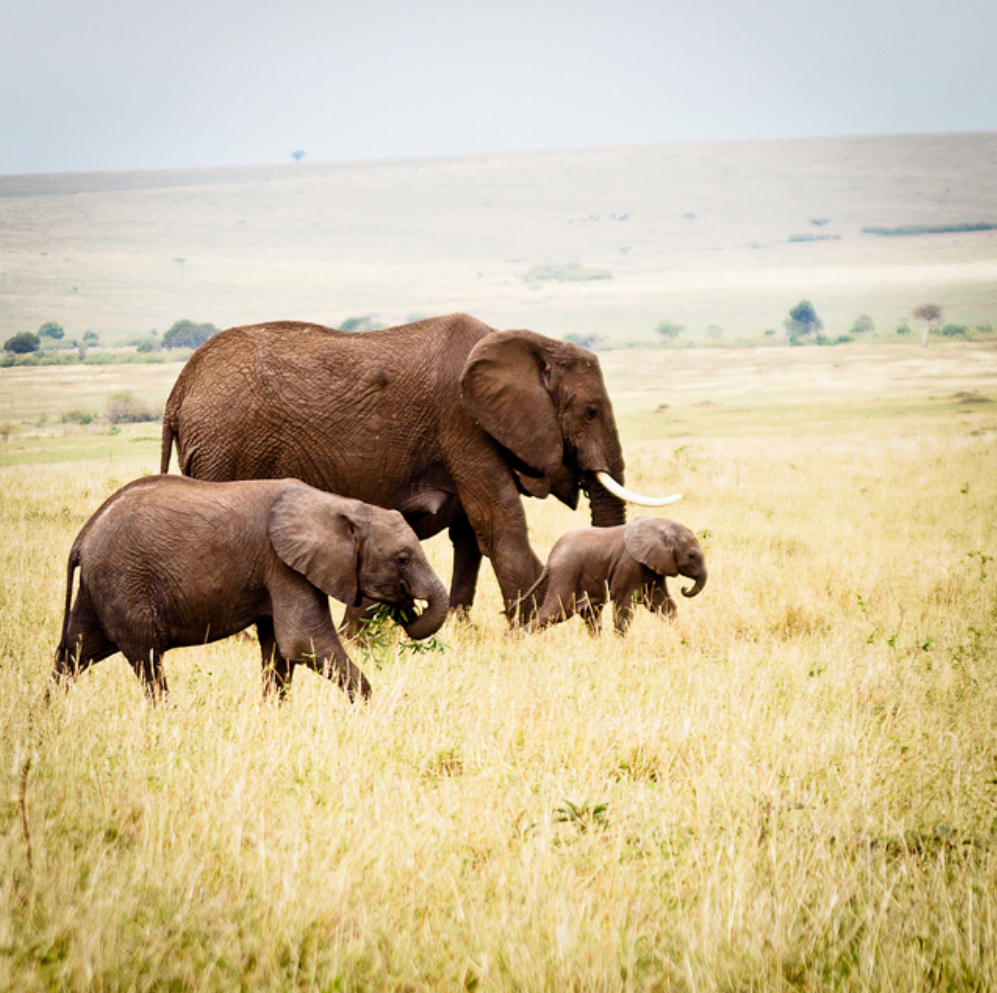} &
            \adjustimage{width=\linewidth,height=5cm,keepaspectratio,valign=m}{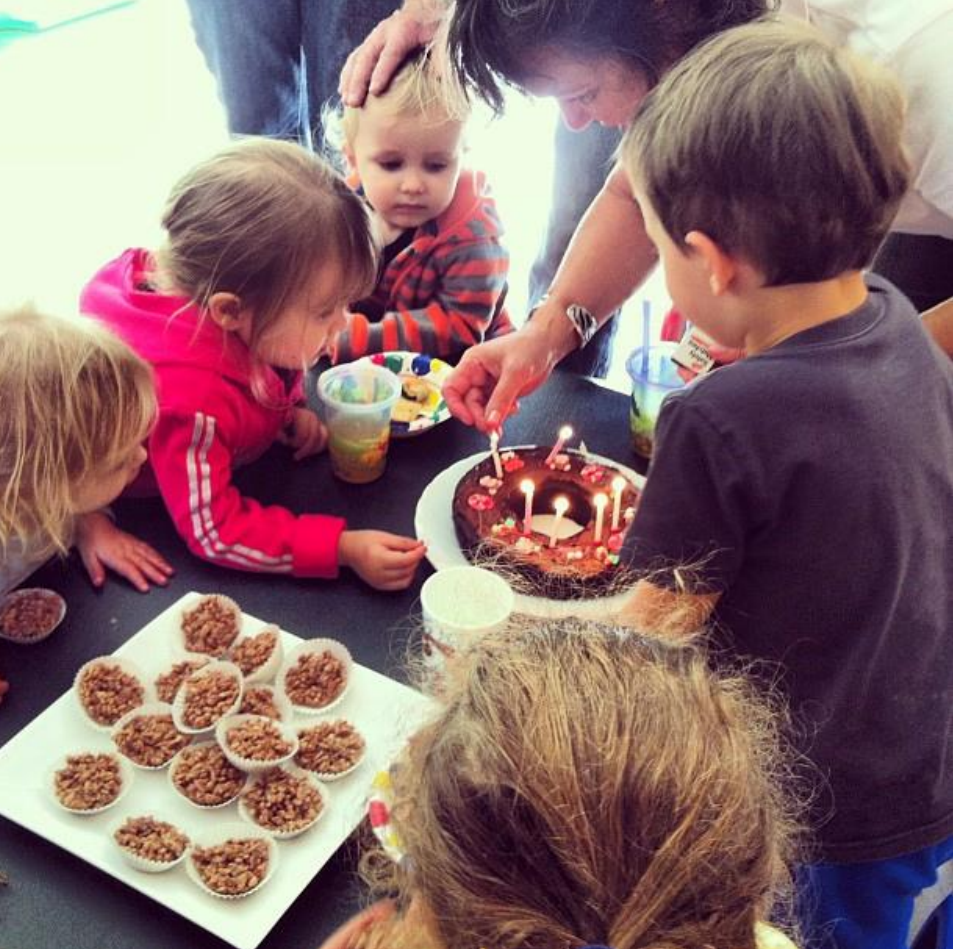} \\
            
            \textbf{Question:} "What color is the Salisbury Rd. sign?" & 
            \textbf{Question:} "How focused is the background?" & 
            \textbf{Question:} "Who is in front of the cake?" \\
            \textbf{Predict answer:} 'blue' & 
            \textbf{Predict answer:} 'not focused' & 
            \textbf{Predict answer:} 'child' \\
            \textbf{Correct answer:} 'white and blue' & 
            \textbf{Correct answer:} 'unfocused' & 
            \textbf{Correct answer:} 'boy' \\

            \adjustimage{width=\linewidth,height=5cm,keepaspectratio,valign=m}{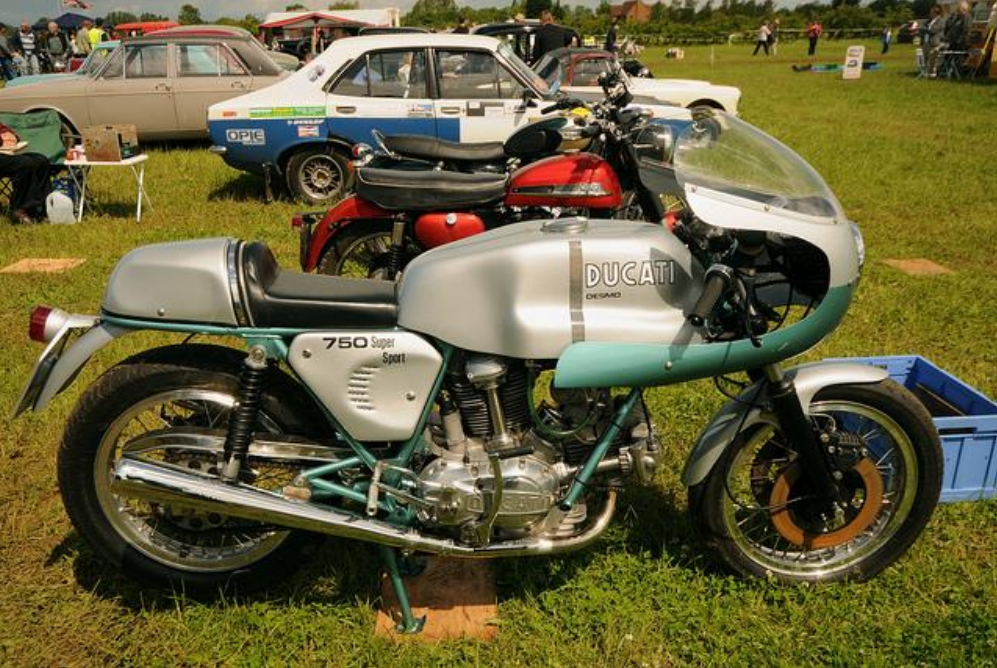} &
            \adjustimage{width=\linewidth,height=5cm,keepaspectratio,valign=m}{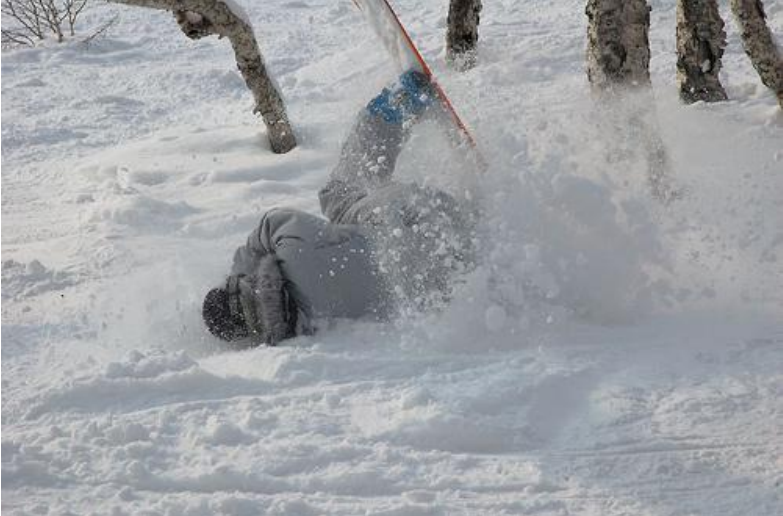} &
            \adjustimage{width=\linewidth,height=5cm,keepaspectratio,valign=m}{} \\
            \textbf{Question:} "Why is there a number on this vehicle?" & 
            \textbf{Question:} "Why is this person leaning down?" & 
            \textbf{Question:} "How many spots on the giraffe?" \\
            \textbf{Predict answer:} 'identification' & 
            \textbf{Predict answer:} 'falling' & 
            \textbf{Predict answer:} 'many' \\
            \textbf{Correct answer:} 'model number' & 
            \textbf{Correct answer:} 'fell' & 
            \textbf{Correct answer:} 'several' \\

            \adjustimage{width=\linewidth,height=5cm,keepaspectratio,valign=m}{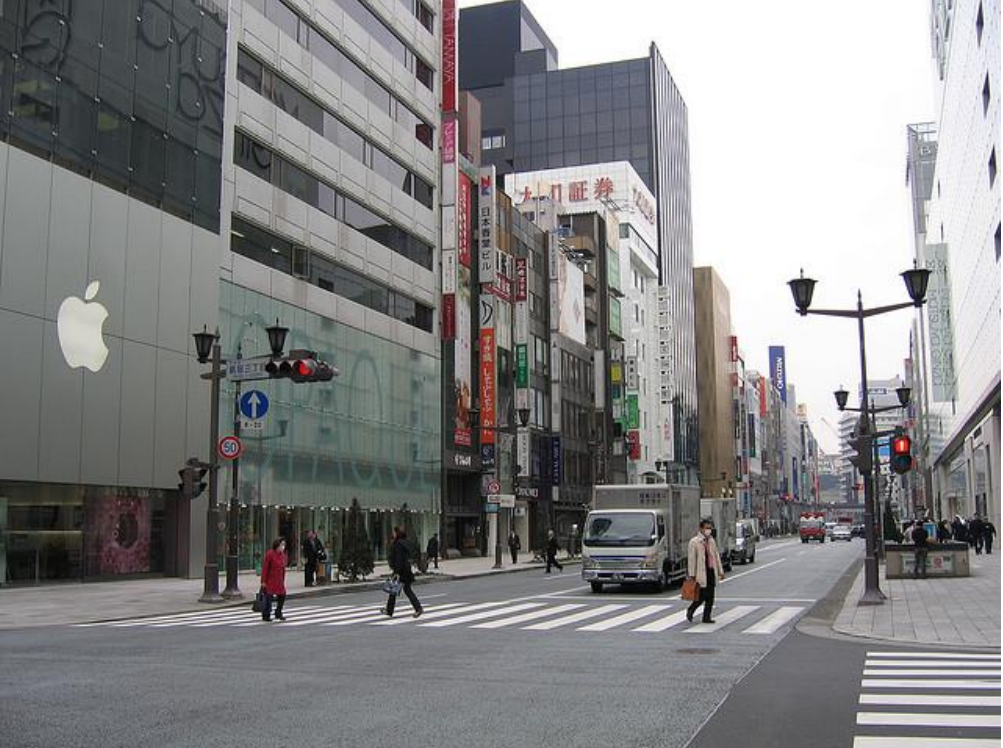} &
            \adjustimage{width=\linewidth,height=5cm,keepaspectratio,valign=m}{} &
            \adjustimage{width=\linewidth,height=5cm,keepaspectratio,valign=m}{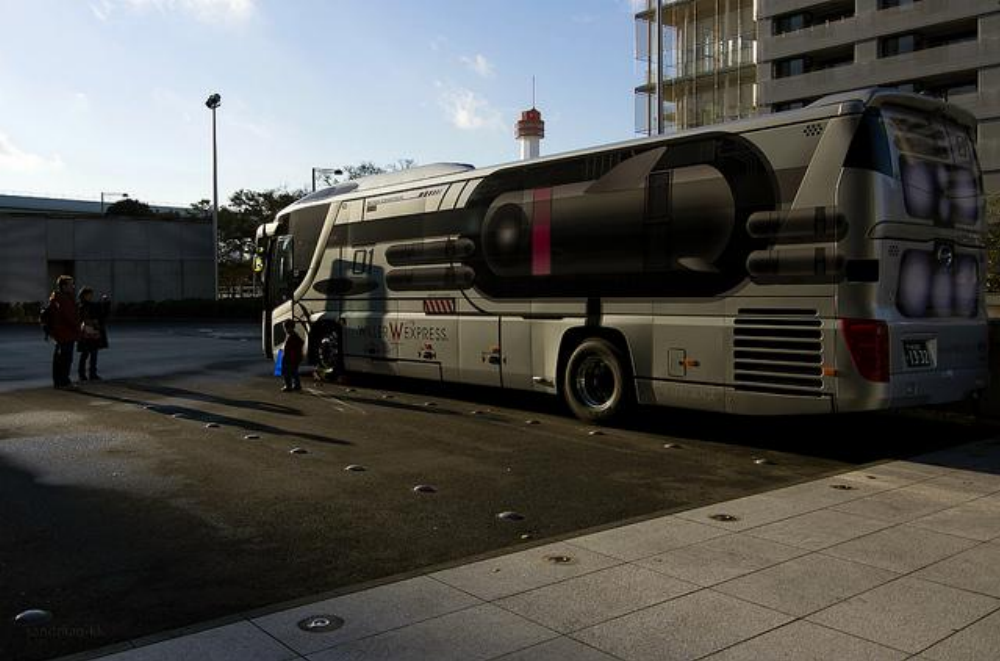} \\
            \textbf{Question:} "Where is the apple emblem?" & 
            \textbf{Question:} "Where would someone find something to dry their hands with in this photo?"
            & 
            \textbf{Question:} "Why does the bus have to stop?" \\
            \textbf{Predict answer:} 'on building' & 
            \textbf{Predict answer:} 'towel' & 
            \textbf{Predict answer:} 'pick up passengers' \\
            \textbf{Correct answer:} 'on building to left' & 
            \textbf{Correct answer:} 'microwave' & 
            \textbf{Correct answer:} 'to load passengers' \\
            
        \end{tabular}
    }
    \caption{Case Study on val2014. This figure illustrates cases where the model’s predicted answers are semantically correct but not counted as correct due to the diversity of possible answers.}
\end{figure*}

\begin{table}[h]
\centering
\renewcommand{\tabcolsep}{16pt} 
\resizebox{\linewidth}{!}{
\begin{tabular}{l|c}
\midrule
\textbf{Input Mode} & \textbf{Accuracy (\%)} \\
\midrule
Speech + Text Mixed Input & 78.85 \\

Pure Text Input & 77.32 \\

Pure Speech Input & 75.62 \\
\midrule
\end{tabular}}
\caption{Model accuracy across different input modes.}
\label{tab:multimodal_robustness}
\end{table}

As shown in \autoref{tab:multimodal_robustness} , the results confirmed that the mixed input mode achieved the highest accuracy, as expected, benefiting from its exposure to both modalities. The pure text input mode performed slightly lower than the mixed input but outperformed pure speech input, reflecting the base model’s strength in text processing. The pure speech input mode showed the lowest accuracy, likely due to the fact that speech understanding was incorporated during fine-tuning rather than being inherent to the base model.

Overall, the model demonstrated stable performance across all input scenarios, with minimal variation in accuracy, indicating that the fine-tuning process effectively enhanced speech understanding while maintaining strong text-processing capabilities.


\subsection{Case Study}


A notable observation from the evaluation is that the model occasionally generates responses that align semantically with ground-truth answers but diverge in phrasing or syntactic structure, leading to misclassification as incorrect. While the core meaning remains consistent, lexical variations or paraphrasing caused automated metrics to flag such responses as errors. This highlights a limitation of rigid evaluation frameworks that prioritize exact textual matches over semantic equivalence. Consequently, the model’s true capability in understanding and reasoning about visual content may be underestimated. Future work should explore more nuanced evaluation protocols, such as incorporating semantic similarity metrics or human judgment, to better capture the model’s functional accuracy.

\section{CONCLUSION}

This work presents SViQA, an end-to-end speech-vision framework that bridges auditory and visual modalities through direct cross-modal alignment, eliminating error-prone text intermediaries while preserving critical acoustic cues for robust question interpretation. By integrating lightweight architecture design and mixed-modal fine-tuning, our approach demonstrates the feasibility of text-free speech-visual fusion, offering enhanced robustness in real-world scenarios compared to cascaded ASR-dependent methods. However, limitations persist: constrained by limited annotated speech-visual datasets and computational resources, the current model adopts a small-scale parameter configuration, potentially restricting its capacity for complex multimodal reasoning. Additionally, the reliance on fixed speech representations may hinder adaptability to diverse acoustic environments. Future work will focus on scalable training strategies and adaptive speech tokenizers to address these challenges, aiming to advance speech-driven multimodal systems toward human-like sensory integration.

\clearpage
\bibliography{anthology,custom}
\bibliographystyle{acl_natbib}




\end{document}